\documentclass[conference]{IEEEtran}
\usepackage{amsmath,graphicx}
\usepackage{algorithmic}
\usepackage{algorithm}
\usepackage{bm}
\usepackage{float}
\usepackage[hidelinks]{hyperref}
\usepackage{multirow}
\usepackage{multicol}

\title{
2D LiDAR and Camera Fusion Using Motion Cues for Indoor Layout Estimation}
\author{\IEEEauthorblockN{Jieyu Li, Robert Stevenson}
\IEEEauthorblockA{Department of Electrical Engineering\\University of Notre Dame\\
Notre Dame, USA \\
jli13@nd.edu, rls@nd.edu}}

\begin{document}
%\ninept
%
\maketitle
\begin{abstract}
This paper presents a novel indoor layout estimation system based on the fusion of 2D LiDAR and intensity camera data. A ground robot explores an indoor space with a single floor and vertical walls, and collects a sequence of intensity images and 2D LiDAR datasets. The LiDAR provides accurate depth information, while the camera captures high-resolution data for semantic interpretation. The alignment of sensor outputs and image segmentation are computed jointly by aligning LiDAR points, as samples of the room contour, to ground-wall boundaries in the images. The alignment problem is decoupled into a top-down view projection and a 2D similarity transformation estimation, which can be solved according to the vertical vanishing point and motion of two sensors. The recursive random sample consensus algorithm is implemented to generate, evaluate and optimize multiple hypotheses with the sequential measurements. The system allows jointly analyzing the geometric interpretation from different sensors without offline calibration. The ambiguity in images for ground-wall boundary extraction is removed with the assistance of LiDAR observations, which improves the accuracy of semantic segmentation. The localization and mapping is refined using the fused data, which enables the system to work reliably in scenes with low texture or low geometric features. 
 %and allows dense reconstruction with planar constraints and complementary information integration for more reliable localization and mapping.
\end{abstract}
\begin{IEEEkeywords}
Indoor layout estimation, multi-view registration, sensor fusion, 2D LiDAR
\end{IEEEkeywords}
\section{Introduction}
Indoor layout estimation is an important problem in robotics, scene understanding and augmented reality. Typical environment perception approaches have generally relied on sensing either range or vision information. This paper proposes a novel indoor layout estimation method based on sensor fusion of a 2D LiDAR and an intensity camera, which allows accurate floor plan reconstruction and semantic segmentation of multiple images for complex indoor environments.

LiDARs, providing accurate depth information by measuring the time-of-flight of laser beams, can be used to develop robust real-time simultaneous localization and mapping (SLAM) systems \cite{zou2021comparative}. But with 3D LiDARs, there is a trade-off between detail and efficiency, and it becomes necessary to extract meaningful higher level features from raw 3D point clouds \cite{droeschel2014local, weingarten20063d, lenac2017fast}. Additional problems occur since LiDARs cannot be tracked reliably in areas with low geometric features (such as long corridors with all parallel walls) and have no observations with glass walls. As a result of these limitations, research has also focused on replacing the 3D LiDAR by the fusion of camera and 2D LiDAR \cite{lu2020Extending, deilamsalehy2016sensor}. For indoor environments under the weak ``Manhattan world'' assumption (vertical walls and a single floor), 2D LiDARs provide a more efficient choice that is still adequate for navigation tasks and polygonal structure modeling as precise maps generated from raw 3D point clouds. 

%Traditional vision-based reconstruction methods, such as structure from motion (SfM) \cite{hartley2003multiple}, can determine depth from camera motion based on the epipolar geometry. However, the estimation is computationally demanding for data association and bundle adjustment, and the resulting structure is commonly sparse and noisy. Furthermore, the method performs poorly in low-texture environments. These problems can be addressed by integrating indoor layout estimation for planar reconstruction \cite{kim2019multi} and point-based SLAM \cite{yang2016pop, bao2011semantic, tsai2011real, tsai2012dynamic}.

For visual-based indoor layout estimation, a common challenge is the detection of ground-wall boundaries. Candidate ground-wall boundaries can be generated based on semantic segmentation and/or geometric assumptions. The strong geometric assumptions, e.g. ``box'' or ``Manhattan world'' assumption, prevent these algorithms from working in complex environments, and image segmentation methods based on visual features \cite{hoiem2005geometric, badrinarayanan2017segnet} are often unreliable when generalized to different environments. With multiple views, the model can be evaluated by the likelihood of feature points in multiple images \cite{tsai2012dynamic}. However, in \cite{tsai2012dynamic}, camera poses are assumed to be known (given by LiDAR data in the experiments), and a larger number of scans is required to estimate and refine the model.

RGB-D sensors can provide both visual and depth information, which become popular in 3D reconstruction of cluttered and small-scale environments \cite{newcombe2011kinectfusion, dai2017scannet}. However, the depth information is noisy and with limited field of view and depth range, so algorithms or even systems integrated with LiDARs \cite{kang2019real, xu2018slam} have been developed to deal with the accumulated errors in tracking. For indoor layout estimation, focusing on polygonal structure modeling, fusion of 2D LiDAR and intensity camera can be a more efficient solution. 

It is a common setup for robots to be equipped with a 2D LiDAR scanning parallel to floors for tracking and a camera capturing vertical and high-resolution data for semantic interpretation \cite{cherubini2014autonomous, liao2017parse}. To jointly analyze the geometric interpretation from different sensors, some systems depend on offline calibration \cite{kwak2011extrinsic, zhou2013new}. However, there are still no commonly used calibration methods in a simple but accurate way for such distinctive sensing modalities. Offline approaches, with manual placement of calibration targets, are not practical solutions for systems where the relative pose between sensors can be changed \cite{debeunne2020review}.
\begin{figure*}[ht]
  \centerline{\includegraphics[trim={0 30 0 20},clip,width=0.95\textwidth]{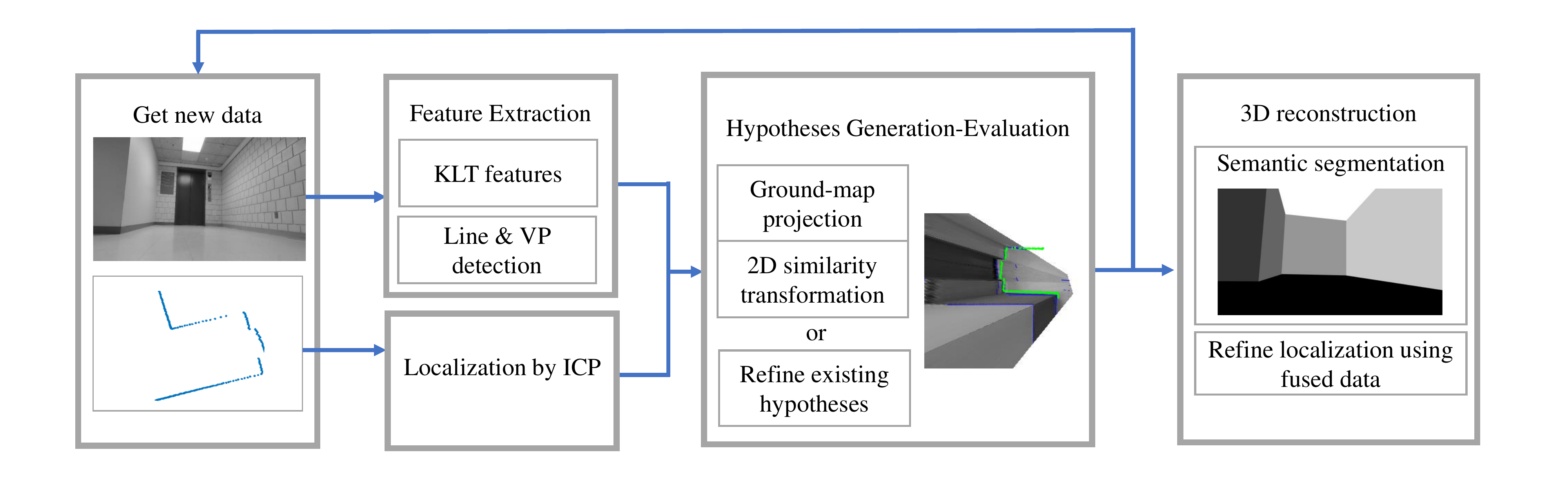}}
%  \vspace{2.0cm}
\label{framework}
\caption{System overview. Features are extracted from images and 2D LiDAR data, and used for generation and evaluation of the hypotheses aligning sensor outputs. Multiple hypotheses are tracked with sequence of data. Finally, the best hypothesis is identified for semantic segmentation and localization refinement.}
\end{figure*}

In the proposed system, the sensor fusion and the layout estimation are performed jointly by projecting LiDAR points to ground-wall boundaries in images. As in our previous work \cite{li2020indoor}, the alignment problem is decoupled into the estimation of a top-down homography and a 2D similarity transformation, and solved by the recursive random sample consensus (rRANSAC) algorithm \cite{niedfeldt2014multiple}. A much more efficient estimation algorithm for the similarity transformation is proposed in this paper based on motion cues between scans, which significantly reduces the number of needed RANSAC iterations from $500$ to $25$. Although the exact relative pose between two sensors is not found, the outputs are aligned in a way useful for indoor layout estimation with geometric constraints (horizontally placed LiDAR and vertical walls). Since the system does not depend on offline extrinsic calibration, if the relative pose between sensors is changed, an automatic re-alignment can be easily implemented with several new scans. In the proposed system, the ambiguity in images for ground-wall boundary extraction is removed with the assistance of LiDAR points. The localization and mapping is further refined with the fused data, which helps the system to work reliably in scenes with low texture or low geometric features.
\section{Proposed System}
\subsection{System Overview}
The proposed system is developed with a 2D LiDAR scanning parallel to ground and an intensity camera with fixed relative position. Data is captured every 10--20 cm along robot trajectories. As shown in Fig. 1, the alignment between LiDAR points and ground-wall boundaries is estimated using an intermediate reference frame that generates the top-down view of images. The rotation between the camera and the reference frame is found according to the vertical vanishing point, and then the 2D similarity transformation between LiDAR points and the top-down--viewed ground-wall boundaries is estimated. The rRANSAC is used to track multiple hypotheses based on sequential measurements by a hypothesis generation-evaluation strategy. When new data is received, features are extracted (Section \ref{datasect}). If the features are consistent with any existing hypotheses, those hypotheses are updated; otherwise, a new hypothesis is generated and stored for further evaluation and optimization. With the aligned data, localization and mapping is refined, and the floor plan is projected onto images for semantic segmentation. The hypothesis generation and the hypothesis evaluation are introduced respectively in Section \ref{hypogen} and Section \ref{evaluatesect}.
%\begin{algorithm}
%\caption{rRANSAC}
%\begin{algorithmic}
%\STATE
%For each new scan:\\
%1. Extract features;\\
%2. Evaluate all existing hypotheses $\bm{\theta}_k\ (k<K)$: for $\bm{\theta}_k$, if $\# inliers >\tau$, update $\bm{\theta}_k$ with inliers from all scans.\\
%3. If no hypotheses is updated, go to step 4; else terminate.\\
%4. Repeat $N$ times to generate a candidate hypothesis $\bm{\theta'}$: get a solution by a randomly selected minimum subset of observations; if more inliers are found, update $\bm{\theta'}$.\\
%5. If $\# inliers >\tau$, refine and store $\bm{\theta'}$ as a new hypothesis $\bm{\theta}_{K+1}$. $K=K+1$ and terminate.

%\end{algorithmic}
%\end{algorithm}

\subsection{Signal Features Detected}
\label{datasect}

As the platform moves on the ground, multiple images and LiDAR scans can be captured. With the LiDAR observation, line features are extracted, and the LiDAR can be tracked by Iterative Closest Point (ICP) algorithm \cite{rusinkiewicz2001efficient}. The floor plan can be reconstructed by integrating multiple scans. 

The image features include a set of Kanade-Lucas-Tomasi (KLT) feature points and straight lines. The KLT feature points of consequent images, corresponding to the same space points, can provide the motion cues. The line segments are grouped into horizontal lines and vertical lines. The horizontal lines are the ground-wall boundary candidates, while the vertical lines are used for the top-down view homography estimation. More details can be found in \cite{li2020indoor}.
\subsection{Hypothesis Generation}
The hypothesis of data alignment is generated using a basic RANSAC algorithm, by estimating a top-down homography and a 2D similarity transformation.
\label{hypogen}
\subsubsection{Top-down View Projection}
\label{top-downsect}
%\begin{figure*}[!thb]
%\centering
%\includegraphics[trim={0 0 0 0},clip,width=0.8\textwidth]{figure/perspectiveimg.pdf}
%\caption{Illustration of an image and its top-down projection in the ``Manhattan world''.}
%\label{top-downview}
%\end{figure*}
The vertical direction in the camera space can be estimated according to the vertical vanishing point as
$\bm{n}_{\rm v}=\frac{1}{h_{\rm v}}\bm{K}^{-1}\tilde{\bm{p}}_{\rm v}$, %-\frac{1}{h_{\rm g}}[x_{\rm g}, y_{\rm g}, f]^{\rm T}=
where $\tilde{\bm{p}}_{\rm v}$ is the homography coordinate of the vanishing point, $\bm{K}$ is the camera intrinsic matrix, and $h_{\rm v}$ is for normalization \cite{caprile1990using}. Let $[x_v,y_v,1]^{\rm T}=\bm{K}^{-1}\tilde{\bm{p}}_{\rm v}$. Given the orthogonality of axes, the top-down reference frame in the camera space can be defined as 
\begin{equation}
\begin{aligned}
    \bm{e}_{\rm X}&= \frac{1}{h_{\rm X}}[-\frac{1}{x_{\rm v}},0,1]^{\rm T} \\
    \bm{e}_{\rm Y}&= \pm\frac{1}{h_{\rm Y}}[x_{\rm v},-\frac{x_{\rm v}^2+1}{y_{\rm v}},1]^{\rm T}, \\
    \bm{e}_{\rm Z}&= \pm\bm{n}_{\rm v}\\
\end{aligned}
\end{equation}
where $h_{\rm X}$ and $h_{\rm Y}$ are the normalization parameters. The sign of $\bm{e}_{\rm Z}$ is chosen so that the direction points to floor independent of the actual camera orientation. The sign of $\bm{e}_{\rm Y}$ makes sure the system follows the right-hand rule. The top-down rotation matrix is $\bm{R}_{\rm g}=[\bm{e}_{\rm X},\bm{e}_{\rm Y},\bm{e}_{\rm Z}]^{\rm T}$, and thus the homography matrix is defined as $\bm{H}_{\rm g}=\bm{K}_{\rm g}\bm{R}_{\rm g}\bm{K}^{-1}$ with $\bm{K}_{\rm g}$ set to be the identity matrix. 

\subsubsection{2D Similarity Transformation Estimation}
\label{2Dsimsect}

The 2D similarity transformation is found to map the points on top-down--viewed ground-wall boundaries to the LiDAR points by $\bm{p}^{\rm l}=\delta\bm{R}_\phi\bm{p}^{\rm g}+\bm{o}^{\rm l}_{\rm g}$, where $\bm{R}_\phi$ is the rotation matrix, $\bm{o}^{\rm l}_{\rm g}$ is the camera center (in XY plane) in the LiDAR coordinate system and $\delta$ is the scaling parameter. In a general case, the minimum subset to determine a 2D similarity transformation, $\bm{p}^{\rm d}=\delta \bm{R} \bm{p}^{\rm s}+\bm{t}$, includes two corresponding point pairs:
\begin{equation}
\begin{aligned}
\delta &=\frac{\sqrt{\Delta {y^{\rm d}}^2+\Delta {x^{\rm d}}^2}}{\sqrt{\Delta {y^{\rm s}}^2+\Delta {x^{\rm s}}^2}}\\
\phi&= {\rm atan2}(\Delta y^{\rm d}, \Delta x^{\rm d})-{\rm atan2}(\Delta y^{\rm s}, \Delta x^{\rm s}),\\
\bm{t}&=\bar{\bm p}^{\rm d}-\delta\bm{R}_\phi\bar{\bm{p}}^{\rm s}
\end{aligned}  
\label{similaritytrans}
\end{equation}
where $\bm{p}^{\rm s}$ and $\bm{p}^{\rm d}$ are the corresponding source and destination points, $\Delta$ indicates the difference (i.e. $\Delta x^{\rm s}=x^{\rm s}_1-x^{\rm s}_2$) and $\bar{\bm{p}}$ indicates the element-wise average value of $\bm{p}$.

The estimation is done by RANSAC that repeatedly selects random minimum subsets to determine model parameters. A straightforward idea is to select two intersections of the ground-wall boundary candidates as the source points and two LiDAR corners as the destination points \cite{li2020indoor}. However, due to the large number of detected horizontal lines and the lack of similarity measure for data association, the probability to select inliers of a plausible solution is low. Thus the maximum number of iterations needs to be large (around 500).

A more effective method is introduced here based on the fact that motion of two sensors can be tracked using their measurements individually and linked to each other by the similarity transformation. For two scans, $i$ and $j$, the transformation of top-down--viewed points can be calculated as 
\begin{equation}
\begin{aligned}
\bm{p}^{{\rm g},j}&=\bm{R}^{\rm l}\bm{p}^{{\rm g},i}+\bm{o}^{{\rm g},j}_{{\rm g},i}\\
&\overset{\rm (a)}{=} \bm{R}^{\rm l}\bm{p}^{{\rm g},i}+\frac{1}{\delta}\bm{R}_{\phi}^{\rm T}(\bm{o}^{{\rm l},j}_{{\rm g},i}-\bm{o}^{\rm l}_{\rm g})\\
&\overset{\rm (b)}{=}\bm{R}^{\rm l}\bm{p}^{{\rm g},i}+\frac{1}{\delta}\bm{R}_{\phi}^{\rm T}({\bm{R}^{\rm l}}\bm{o}^{\rm l}_{\rm g}+\bm{t}^{\rm l}-\bm{o}^{\rm l}_{\rm g}),
\end{aligned}
\label{topdownmotion}
\end{equation}
where $\bm{R}^{\rm l}$ is the rotation matrix of the top-down reference frame, the same as the LiDAR rotation matrix, and $\bm{o}^{{\rm g},j}_{{\rm g},i}$ indicates the $i^{\rm th}$ camera location in the $j^{\rm th}$ top-down reference frame. In ${\rm (a)}$, the top-down--viewed camera location  $\bm{o}^{{\rm g},j}_{{\rm g},i}$ is calculated from the LiDAR-observed camera location $\bm{o}^{{\rm l},j}_{{\rm g},i}$ according to the 2D similarity transformation. ${\rm(b)}$ follows the transformation of LiDAR observation between two scans, with $(\bm{R}^{\rm l}, \bm{t}^{\rm l})$ as the LiDAR motion.

When $\bm{R}^{\rm l}\ne\bm{{\rm I}}_2$, Equation \ref{topdownmotion} can be expressed as
\begin{equation}
    (\bm{{\rm I}}_2-{\bm{R}^{\rm l}})^{-1}(\bm{p}^{{\rm g},j}-{\bm{R}^{\rm l}} \bm{p}^{{\rm g},i})=\frac{1}{\delta}\bm{R}_{\phi}^{\rm T}[(\bm{{\rm I}}_2-{\bm{R}^{\rm l}})^{-1}\bm{t}^{\rm l}-\bm{o}^{\rm l}_{\rm g}]. 
\label{method2}
\end{equation}
Thus Equation \ref{similaritytrans} can be solved with
\begin{equation}
\begin{aligned}
\bm{p}^{\rm d}&=(\bm{{\rm I}}_2-{\bm{R}^{\rm l}})^{-1}\bm{t}^{\rm l} \\
\bm{p}^{\rm s}&=(\bm{{\rm I}}_2-{\bm{R}^{\rm l}})^{-1}(\bm{p}^{{\rm g},j}-{\bm{R}^{\rm l}} \bm{p}^{{\rm g},i}),
\end{aligned}
\end{equation}
where $\bm{p}^{{\rm g,i}}$ and $\bm{p}^{{\rm g,j}}$ are the top-down view of tracked KLT features ${\bm{p}}^{{\rm c}}$, calculated by $\bm{p}^{{\rm g}}=[\frac{\bm{h}_{\rm g1}\tilde{\bm{p}}^{{\rm c}}}{\bm{h}_{\rm g3}\tilde{\bm{p}}^{{\rm c}}},\frac{\bm{h}_{\rm g2}\tilde{\bm{p}}^{{\rm c}}}{\bm{h}_{\rm g3}\tilde{\bm{p}}^{{\rm c}}}]^{\rm T}$ with $\bm{h}_{{\rm g}n}$ as the $n^{\rm th}$ $(n=1,2,3)$ row of the top-down homography matrix. Since $\bm{p}^d$ depends solely on the LiDAR motion, three scans are required to generate the minimum subset including two different point pairs. 

Whether the selected point pairs are inliers depends on (1) if they are associated correctly to the same space point and (2) if the space point is on ground. The first condition can be mostly ensured by the KLT tracker. The chance of the second condition can be increased by giving higher weights for points on the bottom part of images. Using this method, the maximum number of iterations is reduced from $500$ to $25$.
%when $\bm{R}^{\rm l}_{i \rightarrow j}\ne\bm{{\rm I}}_2$ can work too.

\subsection{Hypothesis Evaluation and Optimization}
\label{evaluatesect}
The hypotheses are evaluated according to the epipolar constraint that describes the geometric relations between projection of a space point from different perspectives as $\tilde{\bm{p}}'^{\rm{T}}\bm{F}\tilde{\bm{p}}=0$. 
The fundamental matrix $\bm{F}$ is calculated by
\begin{equation}
    \bm{F}=\bm{K}'^{-\rm T}[\bm{t}^{\rm c}]_{\times}\bm{R}^{\rm c}\bm{K}^{-1},
\end{equation}
where $(\bm{R}^{\rm c},\bm{t}^{\rm c})$ is the relative camera pose, and $[\bm{t}^{\rm c}]_{\times}$ is the matrix representation of the cross product with $\bm{t}^{\rm c}$. Given the hypothesis, the camera motion can be calculated by
\begin{equation}
    \begin{aligned}
\bm{t}^{\rm c}&=\frac{1}{\delta}\bm{R}_{\rm g}^{\rm T}\bm{R}_{\phi}^{\rm T}({\bm{R}^{\rm l}}\bm{o}^{\rm l}_{\rm g}+\bm{t}^{\rm l}-\bm{o}^{\rm l}_{\rm g}) \\
\bm{R}^{\rm c}&=\bm{R}_{\rm g}^{\rm T}\bm{R}^{\rm l}\bm{R}_{\rm g},    
    \end{aligned}
\end{equation}
where 2D rotation matrices ($\bm{R}^{\rm l}$ and $\bm{R}_\phi$) and 2D translation ($\bm{t}^{\rm l}$ and $\bm{o}^{\rm l}_{\rm g}$) are extended to 3D space.

The inliers are the KLT feature pairs that satisfy the epipolar constraints, as $\delta^2(\tilde{\bm{p}}'^{\rm{T}}\bm{F}\tilde{\bm{p}})^2<\tau$. Since the visual-based reconstruction has scale ambiguity, the effect of $\delta$ needs to be eliminated. The hypothesis with most inliers is identified as the solution, which projects the LiDAR points around the ground-wall boundaries. The ground-wall boundary identification and hypothesis optimization can be done by minimizing point-to-line metrics.

\section{Experiments and Results}
\label{resultsec}

\begin{figure}[htb]
  \centerline{\includegraphics[trim={20 100 20 40},clip,width=8.5cm]{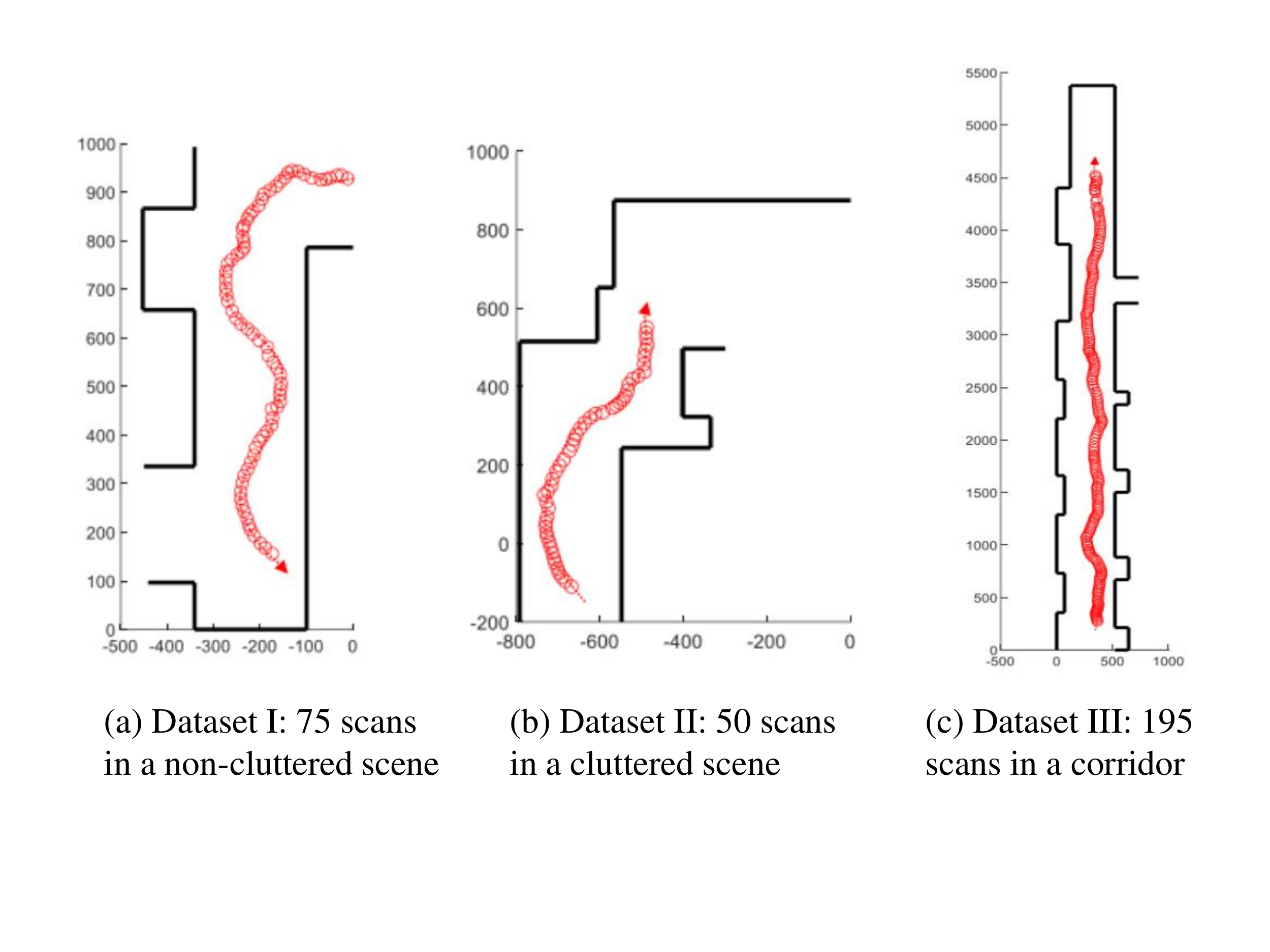}}
%  \vspace{2.0cm}
\caption{Floor plans and robot trajectories (in cm).}
\label{datasetfloorplan}
\end{figure}

The proposed approach is tested using three datasets captured at the University of Notre Dame with different environments, a non-cluttered area (Dataset I), a cluttered area (Dataset II) and a long corridor (Dataset III)\footnote{The datasets can be found by the link: \url{https://github.com/leejieyu13/LidarCameraDataset}}. The floor plans and robot trajectories are shown in Fig. \ref{datasetfloorplan}. The images, with resolution $1920\times1080$, were collected by a calibrated camera with unknown poses, while the LiDAR data was captured by Slamtec's ``RPLIDAR A3.'' The ground truth for semantic segmentation are generated by manually labeling the planes (i.e. the walls and the ground plane). 

To the best of our knowledge, the similar camera and 2D LiDAR setup for indoor layout estimation systems, and thus directly comparable work, have not been found. The method is compared with a visual-based planar SLAM approach, Pop-up SLAM \cite{yang2016pop}, to demonstrate the advantages brought by sensor fusion. Similar to the proposed system, Pop-up SLAM also combines the scene understanding and plane detection for localization and mapping, which shows advantages especially in low-texture environments. In Pop-up SLAM, ground-wall boundaries are selected according to semantic segmentation by SegNet \cite{badrinarayanan2017segnet}, and the planar SLAM is solved by Incremental Smoothing and Mapping \cite{kaess2008isam}. For a more fair comparison, results from LiDAR-based SLAM are provided to the Pop-up SLAM as odometry data. For semantic segmentation, the reconstructed floor plan is projected onto the image according to the camera poses. The performance of both floor plan reconstruction and semantic segmentation is evaluated.

\begin{table*}[htb]
\begin{center}
% \resizebox{\textwidth}{!}{
\begin{tabular}{ c c c c c c c c c c }
\hline
%  \multirow{2}{}{}& 
 &\multicolumn{3}{c}{non-cluttered} & \multicolumn{3}{c}{cluttered} &\multicolumn{3}{c}{\centering long corridor} \\
 &segmentation & RMSE & F-score &segmentation & RMSE & F-score &segmentation & RMSE & F-score \\
\hline
Proposed &97.2\% & 0.10m & 97.8\% &93.8\% & 0.21m & 96.2\% &95.5\% &0.48m &95.3\%\\
\hline
%seg 94.8%, rmse = 0.98m, f = 91.3%
\cite{yang2016pop} & 96.9\% & 0.50m  & 95.1\% & 
   86.4\% & 0.39m & 90.9\%  & 92.2\% & 2.19m & 87.5\%\\
\hline
\end{tabular}
% }
\end{center}
\caption{Quantitative comparison.}
\label{result}
\end{table*}

\begin{figure*} [!thb]
\centering
\includegraphics[trim={0 185 0 0},clip, width=0.9\textwidth]{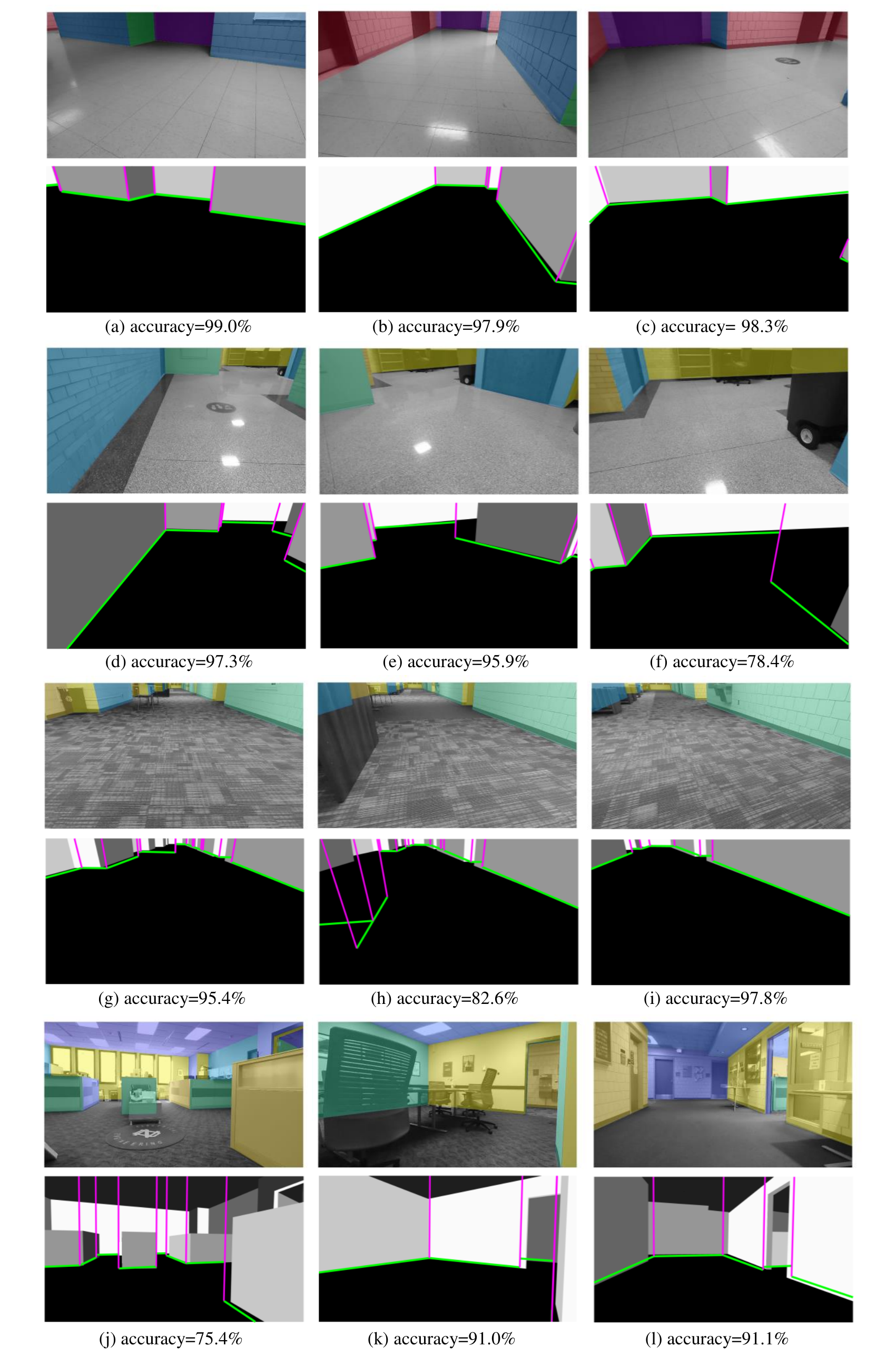}
\caption{Experimental results: the upper rows are the image data and semantic segmentation ground truth, and the lower rows are the segmentation results with wall edges (magenta lines) and ground-wall boundaries (green lines) compared with the ground truth labels. (a-c) for Dataset I,  (d-f) for Dataset II and (g-i) for Dataset  III.}
\label{dataresult}
\end{figure*}

\begin{figure*}
  \centerline{\includegraphics[trim={20 11 20 11},clip,width=0.95\textwidth]{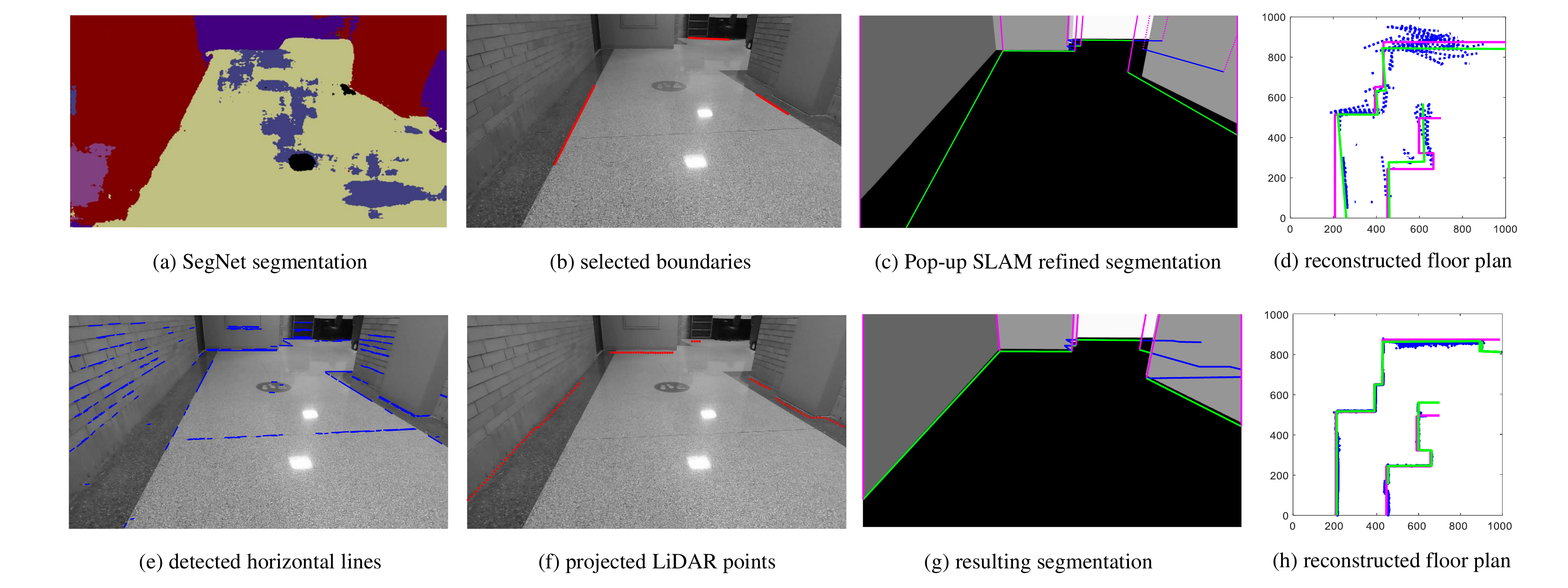}}
%  \vspace{2.0cm}
\caption{Examples of Pop-up SLAM (upper row) and the proposed system (lower row): the first two columns show the approaches for ground-wall boundary identification; the third column is the segmentation results with visible boundaries (green), occluded boundaries (blue) and edges of walls (magenta), compared to the ground truth (indicated by different gray levels); the last column shows the reconstructed floor plans (green), ground truth (magenta) and transferred observations (blue).}
\label{badpopup}
\end{figure*}

\begin{figure*}
\centering
\includegraphics[trim={10 0 10 0},clip, width=\textwidth]{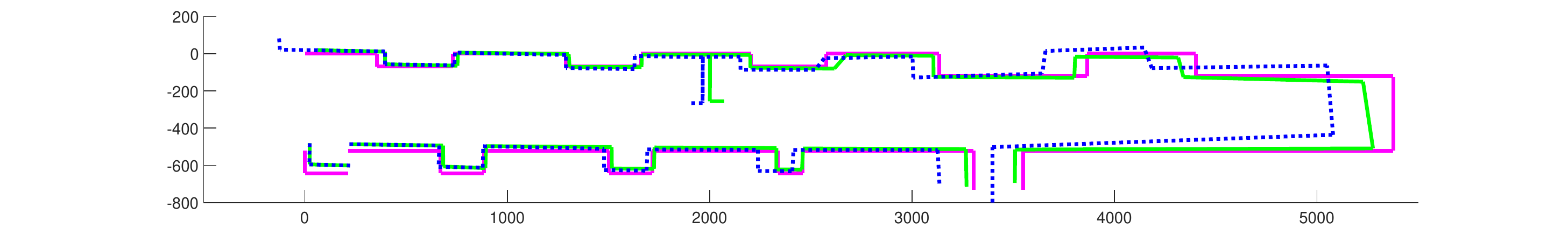}
\caption{Comparison of floor plan reconstruction of Dataset III. The floor plan reconstructed by LiDAR-based SLAM and the proposed fusion SLAM are illustrated respectively by blue dash lines and green lines. The ground truth is indicated by magenta lines.}
\label{slamcompare}
\end{figure*}
The results can be found in Fig. \ref{dataresult} and Table \ref{result}. The semantic segmentation is evaluated by the classification accuracy as the percentage of correctly labeled pixels. The quality of the floor plan reconstruction is evaluated by root mean square error (RMSE) of the wall intersections and the F-score. The F-score is the ratio of overlapping area to the mean of reconstructed and true area, $\mathcal{F}=\frac{2|S_{\rm gen}\cap S_{\rm tru}|}{|S_{\rm gen}| + |S_{\rm tru}|}$, which indicates the similarity of reconstructed shape and ground truth. The results show that the system works well in various environments. Some errors are mainly introduced by large furniture in the cluttered scene that cannot be distinguished from walls based on LiDAR data. 

The comparison to \cite{yang2016pop} shows that the proposed method can achieve better accuracy by combining LiDAR data and image sequences. Fig. \ref{badpopup} shows an example of comparison to further demonstrate the performance improvement. The proposed method identifies the ground-wall boundaries in images by the projected LiDAR points (Fig. \ref{badpopup}(f)). The endpoints of walls can also be easily determined by the LiDAR observation from multiple scans with wider field of view. The Pop-up SLAM depends on semantic segmentation of single images, which is sometimes unreliable, especially with poor lighting and patterns or reflections/shadows on floors (Fig. \ref{badpopup}(a)). It causes wrong selection of ground-wall boundaries, as shown in Fig. \ref{badpopup}(b). Furthermore, even when the ground-wall boundaries are selected correctly, the lines far away from camera have large observation noise. As shown in Fig. \ref{badpopup}(d), the transferred boundaries (blue dash lines) have larger variance compared to the LiDAR observation (blue dots in Fig. \ref{badpopup}(h)). Although with the assistance of the LiDAR-based localization in the experiments, the image noise challenges the data association and propagates to mapping. To improve the performance, heavy optimization with more overlapping observations is needed.

Furthermore, the fused data can be used to refine the localization and mapping. Although the LiDAR provides accurate spatial data, the LiDAR-based localization problem might be underdetermined in the scenes with low geometric features (when all the detected walls are parallel to each other as in long corridors). This can be solved by tracking both LiDAR points and KLT features in images. Fig. \ref{slamcompare} shows the improvement of the floor plan reconstruction with fused data for Dataset III. The RMSE of landmarks (wall intersections) is reduced from $0.98$m to $0.48$m, and the F-score is increased from $91.3\%$ to $95.3\%$ for the floor plan reconstruction.

%\begin{figure}[htb]
%  \centerline{\includegraphics[trim={0 20 0 10},clip,width=8.5cm]{figure/failurecase.pdf}}
%  \vspace{2.0cm}
%\caption{Failure case: raw image (left) and the semantic segmentation (right) with accuracy = 82.3\%.}
%\label{failurecase}
%\end{figure}
%Fig. \ref{failurecase} shows a failure case of the proposed system. Although the results show that the system works well in various environments, the errors are mainly introduced by large clutter, as the table in this example which cannot be distinguished from walls based on LiDAR data.

\section{Conclusion}
\label{conclusion}
The paper presents a complete system for indoor layout estimation based on the fusion of 2D LiDAR and intensity camera. Without offline extrinsic calibration, sensor outputs are aligned by projecting the LiDAR points, as sampled from the room contour, to the ground-wall boundaries.

Comparison between a visual-based planar reconstruction method \cite{yang2016pop} shows that the LiDAR helps to remove the ambiguity in images for ground-wall boundary extraction. The fused data is used to improve the accuracy and the robustness of multi-view registration. The system achieves accurate floor plan reconstruction and semantic segmentation in various environments. The proposed method avoids the need of strong layout assumptions or heavy computation with many overlapping images, and allows the multi-view registration to work reliably in scenes with low texture or low geometric features. The complementary information can be further used for object detection to develop an improved system working in complex environments with clutter.

\bibliographystyle{IEEEbib}
\bibliography{ref}

\end{document}